\documentclass[sigconf,nonacm]{acmart}

\usepackage{amsmath,booktabs,array,multirow,makecell,tikz}
\usetikzlibrary{arrows.meta,positioning,fit}
\newcommand{\model}{AGT}

\title{Accounting Graph Transformer for Short-History Multi-KPI Forecasting in Small Businesses}

\author{Shrutendra Harsola}
\affiliation{%
  \institution{Foresight-AI, Intuit}
  \city{Bangalore}
  \country{India}
}

\author{Vignesh Subrahmaniam}
\affiliation{%
  \institution{Foresight-AI, Intuit}
  \city{Bangalore}
  \country{India}
}

\begin{abstract}
Small businesses often have only 12--24 months of accounting history, yet planning and risk workflows require coordinated forecasts across financial statements. We study joint 12-month forecasting of 13 income-statement, balance-sheet, cash-flow, and working-capital key performance indicators (KPIs) from 71 monthly ledger series. We introduce the Accounting Graph Transformer (\model), which represents each ledger series as a masked token, exchanges information through typed attention on a fixed accounting-relation graph, pools target-specific context, and fuses it with a gated three-month recency path. Across 11,993 forecast origins from 1,060 unseen companies, \model{} achieves sample-weighted KPI-macro mean absolute error (MAE) $0.6990\pm0.0013$ over three independent seeds, compared with $0.7378\pm0.0014$ for the strongest baseline, LightGBM. At the pre-specified seed 42, a paired company-clustered bootstrap gives a LightGBM-minus-\model{} difference of 0.0395 with 95\% confidence interval (CI) $[0.0350,0.0439]$. \model{} is best on all 13 KPIs against LightGBM, TimeMixer, and SOFTS in the matched seed-42 comparison, while final-architecture ablations show that relational attention, accounting topology, and the recency path each improve validation and test accuracy. On 7,094 additional unseen companies with origins sampled from January--May 2025, \model{} obtains 0.7548 MAE versus 0.7694 for SOFTS. A single 5.3M-parameter model produces 156 aligned forecasts without company-specific fitting, providing one forecasting layer for integrated planning, liquidity, and working-capital analysis.
\end{abstract}


\begin{document}
\maketitle

\section{Introduction}
Financial planning for a small business rarely depends on one series in isolation. Revenue projections affect expected receivables and operating cash flow; inventory and cost behavior shape payables; asset growth changes the financing required to keep the balance sheet in balance. A useful forecasting system should therefore produce a joint set of trajectories for income-statement, balance-sheet, cash-flow, and working-capital quantities rather than forecast revenue alone.

This joint view is valuable in several finance workflows. A liquidity review needs revenue, operating expenses, receivables, payables, and cash-flow forecasts on the same horizon. A budgeting workflow must connect expected sales to the asset and financing requirements that support them. Portfolio monitoring benefits from one model that can score many businesses without maintaining a separate forecaster for each firm. These uses place a premium on cross-company generalization, stable behavior with sparse inputs, and a forecast representation that can exploit financial structure.

The setting is difficult for standard multivariate forecasting models. Small businesses commonly have only one to two years of monthly history. Their charts of accounts are heterogeneous, many subaccounts are inactive, and the same company may contribute several overlapping forecast origins. At the same time, the channels are not exchangeable. Accounting identities and operational links specify a sparse pattern of plausible dependencies: revenue is related to accounts receivable, cost of goods sold (COGS) and expenses to accounts payable, the asset groups to liabilities and equity, and the three cash-flow sections to one another. These relations provide an external structural prior that does not need to be inferred from a long company-specific sequence.

We formulate a company-disjoint panel task with 71 monthly ledger series, 13 target KPIs, 12--24 observed months, and a 12-month horizon. The schema combines top-level KPIs, company-specific ranked subaccounts, catch-all accounts, and accounts-receivable/accounts-payable (AR/AP) aging buckets. The proposed Accounting Graph Transformer (\model) maps each series to a masked token, applies four relational-attention blocks over a fixed accounting graph, pools a KPI-specific representation, and combines it with the last three observations of that KPI. The graph restricts cross-series information flow to accounting and accrual relations, while the recency path preserves local target dynamics that are easy to dilute in a joint encoder.

This paper makes three contributions. First, it formulates and evaluates a large, company-disjoint forecasting task across all three financial statements and working capital, with explicit treatment of short histories, missing accounts, and repeated origins. Second, it introduces a compact architecture that combines a fixed accounting-relation graph with target-specific recency conditioning. Third, it evaluates the model against statistical, tree, neural, and time-series foundation models under a common temporal-availability, masking, target, and scoring pipeline. The evaluation includes matched independent seeds, company-clustered inference, per-KPI results, final-architecture ablations, company-balanced estimates, and a later-origin cohort of unseen firms.

The evidence is broad. \model{} improves on the strongest tree and generic neural baselines, wins on every individual KPI in the matched comparison, and remains ahead when each company receives equal weight. The final-architecture ablations identify both relational attention and the recency path as useful: deleting graph attention produces the largest degradation, while a random graph is also worse than the accounting topology. These findings establish accounting structure as a useful inductive bias for short company panels.

\section{Related Work}
Long-horizon neural forecasters include PatchTST~\cite{nie2023patchtst}, iTransformer~\cite{liu2024itransformer}, TimeMixer~\cite{wang2024timemixer}, TiDE~\cite{das2023tide}, DLinear~\cite{zeng2023linear}, and SOFTS~\cite{han2024softs}. They provide strong generic sequence backbones but typically learn channel interactions without an external financial schema. Chronos-2~\cite{ansari2025chronos2}, TimesFM~\cite{das2024timesfm}, and Moirai 2.0~\cite{liu2025moirai2} broaden the pretraining distribution and test whether general time-series pretraining transfers to sparse company ledgers.

Global forecasting fits one shared model across a panel rather than a separate model for every series, often improving generalization when individual histories are short~\cite{montero2021global}. AGT is global across companies and multivariate within each company: parameters are shared across the panel, while schema embeddings and masks adapt the representation to each firm's active accounts and observed history.

Graph forecasting models commonly infer adjacency or inter-series relationships from data~\cite{wu2020connecting,shang2021discrete,bai2020adaptive,cao2020spectral}. \model{} instead tests an exogenous accounting graph that is fixed before model fitting and shared by every company, including firms unseen during training. Recent relational transformer work also injects graph structure into forecasting backbones~\cite{graft2026}. Our graph connects series that share a statement identity, a parent account, or an operational accrual relationship; the ablations compare this fixed topology with graph removal and degree-matched random connectivity. Unlike a fully connected multivariate encoder, it does not ask every ledger series to interact with every other series at every layer.

Accounting-variable forecasting has previously used component hierarchies~\cite{qiao2018hierarchical}. Our task extends this direction to joint forecasting of 13 KPIs across all three statements from 71 ledger series and evaluates the resulting model on entirely unseen companies. The account schema is also heterogeneous across businesses: child slots are ranked within a parent rather than defined by a universal chart of accounts, so the model must combine stable statement-level meaning with company-specific account composition.

Forecast reconciliation is related but distinct. Reconciliation adjusts independently produced forecasts to satisfy grouped constraints~\cite{wickramasuriya2019optimal,taieb2017coherent}, whereas \model{} uses accounting relations inside the representation learner. The two operations address different questions: relational attention determines which histories inform a forecast, while reconciliation determines whether a completed forecast obeys a chosen identity. The present work evaluates the predictive role of accounting relations before any downstream projection.

\section{Task and Data}
\subsection{Input schema and preprocessing}
For company $c$ and cutoff month $t$, let the forecast origin be $o=(c,t)$. The input is $X_o\in\mathbb{R}^{71\times24}$; a temporal mask $T_o\in\{0,1\}^{71\times24}$ marks observed months and a series mask $m_o\in\{0,1\}^{71}$ marks available ledger series. The 13 top-level series are revenue, COGS, expense, current assets, fixed assets, other assets, liabilities, equity, operating cash flow, investing cash flow, financing cash flow, accounts receivable, and accounts payable.

The remaining 58 series describe account composition. Each income-statement parent uses the five largest subaccounts for that company, ranked by total absolute pre-origin activity, plus a catch-all that aggregates the remainder. Balance-sheet and cash-flow parents use the three largest subaccounts plus a catch-all. AR and AP each use four aging buckets. Selection is recomputed at each forecast origin using only the observed window. A ranked child slot therefore represents a parent and within-parent activity rank rather than a globally identical account name. The learned slot embedding identifies both pieces of information. COGS, expense, liabilities, and equity are stored as positive magnitudes; cash-flow components retain signed net values.

The parent value and its children are all retained as separate input channels. This preserves the directly observed KPI while exposing changes in account composition. A catch-all prevents the representation from discarding the long tail of company-specific accounts, and the fixed number of slots permits batching across heterogeneous charts of accounts.

\begin{table}[t]
\centering
\caption{Composition of the fixed 71-series input schema. Child-slot counts include the catch-all; AR/AP children are aging buckets.}
\label{tab:schema}
\setlength{\tabcolsep}{3.0pt}
\small
\begin{tabular}{lrrr}
\toprule
Block & Parents & Slots/parent & Series \\
\midrule
Top-level KPIs & 13 & -- & 13 \\
Income-statement children & 3 & 6 & 18 \\
Balance-sheet children & 5 & 4 & 20 \\
Cash-flow children & 3 & 4 & 12 \\
AR/AP aging buckets & 2 & 4 & 8 \\
\midrule
Total & & & 71 \\
\bottomrule
\end{tabular}
\end{table}

Each history is represented in a 24-month tensor, with unobserved leading months marked by the temporal mask. Both normalization stages use only observed months, and masked positions are set to zero before projection. A genuine zero within an active series remains an observed value; only a series with no observed activity is marked unavailable. The temporal mask is therefore distinct from the series-availability mask: the first records company age, while the second records whether a ledger channel is active at all.

A shared cache scales each observed series by its trailing mean absolute activity. \model{} then z-normalizes each series over observed months before token projection, analogous to Reversible Instance Normalization (RevIN)~\cite{kim2022revin}. The resulting tokens represent standardized trajectory shape together with schema identity; the trailing mean in Eq.~\eqref{eq:target} remains available to convert relative forecasts back to dollar values. All preprocessing statistics and scoring percentiles are fit on the training split only.

\subsection{Forecast target and estimands}
For KPI $k$ and horizon $h$ at origin $o=(c,t)$, the mean-relative target is
\begin{equation}
 y_{o,k,h}=\frac{v_{c,k,t+h}-\mu_{o,k}}{|\mu_{o,k}|},\qquad h=1,\ldots,12,
 \label{eq:target}
\end{equation}
where $v_{c,k,t+h}$ is the future dollar value, $\mu_{o,k}$ is the signed mean of KPI $k$ over the observed trailing 12 months, and $\hat y_{o,k,h}$ denotes the model forecast of $y_{o,k,h}$. Thus $y=0$ predicts the trailing mean, while $y=\pm1$ is one trailing-mean magnitude above or below it. Sample--KPI pairs with $|\mu_{o,k}|<10^{-6}$ are treated as inactive and excluded for that KPI only; this affects fewer than 3\% of pairs.

AGT, the generic neural baselines, and LightGBM use the same target-construction pipeline during fitting. For scoring, let $\ell_k$ and $u_k$ be the KPI-specific 2.5th and 97.5th training percentiles and define $\bar y=\min(u_k,\max(\ell_k,y))$, with the same transformation applied to $\hat y$. The error for origin $o$ and KPI $k$ is
\begin{equation}
 e_{o,k}=\frac{1}{12}\sum_{h=1}^{12}|\bar y_{o,k,h}-\bar{\hat y}_{o,k,h}|.
 \label{eq:originerror}
\end{equation}
The headline sample-weighted KPI-macro MAE averages $e_{o,k}$ over origins within each KPI and then equally over the 13 KPIs. Paired uncertainty uses a company-clustered bootstrap that resamples companies and moves all their origins together. We also report
\begin{equation}
\operatorname{MAE}_{\mathrm{company}}=
\frac{1}{13}\sum_{k=1}^{13}\frac{1}{|C_k|}\sum_{c\in C_k}
\frac{1}{|O_{c,k}|}\sum_{o\in O_{c,k}}e_{o,k},
\label{eq:companymae}
\end{equation}
where $C_k$ is the set of companies with at least one eligible origin for KPI $k$, and $O_{c,k}$ is company $c$'s corresponding origin set. This company-balanced estimate gives every eligible firm equal weight. Paired comparisons use seed 42, fixed before test evaluation; across-seed mean and standard deviation are reported separately for the four leading trainable methods.

\subsection{Accounting relations}
The top-level financial relations comprise three statement groups and four accrual links. At the top level, the balance-sheet variables satisfy
\begin{equation}
\underbrace{v_{\mathrm{CA}}+v_{\mathrm{FA}}+v_{\mathrm{OA}}}_{\mathrm{Assets}}
= v_{\mathrm{L}}+v_{\mathrm{E}},
\label{eq:bs}
\end{equation}
where $v_{\cdot}$ denotes a monthly dollar value at a common company-month, and CA, FA, OA, L, and E denote current assets, fixed assets, other assets, liabilities, and equity. The profit-and-loss (P\&L) group is organized around
\begin{equation}
\mathrm{NetIncome}=v_{\mathrm{Revenue}}-v_{\mathrm{COGS}}-v_{\mathrm{Expense}},
\label{eq:pnl}
\end{equation}
and the cash-flow group around
\begin{equation}
\Delta\mathrm{Cash}=v_{\mathrm{OCF}}+v_{\mathrm{ICF}}+v_{\mathrm{FCF}}.
\label{eq:cf}
\end{equation}
Here OCF, ICF, and FCF denote operating, investing, and financing cash flow. Net income, total assets, and change in cash are derived quantities rather than forecast targets. AGT therefore uses Eqs.~\eqref{eq:bs}--\eqref{eq:cf} to define neighborhoods among the observed terms instead of adding synthetic output nodes. The graph also contains revenue--AR, COGS--AP, expense--AP, and operating-cash-flow--equity pairs. These links encode recurring operational relationships while leaving their strength to the learned relation gate.

The graph specifies which histories may exchange information, while the learned relation gate determines the strength of each connection for each destination node. Dollar forecasts are recovered after prediction as
\begin{equation}
\hat v_{c,k,t+h}=\mu_{o,k}+|\mu_{o,k}|\hat y_{o,k,h},\qquad o=(c,t).
\label{eq:invert}
\end{equation}
This separation lets the encoder operate on a cross-company relative scale while retaining a direct mapping back to business units.

\subsection{Cohorts}
The data contain anonymized monthly general-ledger aggregates from a commercial cloud-accounting platform. Companies are split before origin construction, so a business appears in exactly one of train, validation, or test. All three primary cohorts use the same 13 monthly forecast origins, from December 2023 through December 2024. Origins require active trailing revenue ($|\mu_{o,\mathrm{revenue}}|\geq10^{-6}$) and a fully observed 12-month horizon; this rule is common to all methods. The resulting test set contains 11,993 forecast origins from 1,060 companies; the average company contributes about 11 origins and the median contributes 12. Mean observed history is 18.4 months (standard deviation 3.5), and 92.2\% of test origins have fewer than 24 observed months. The forecast horizon is fully observed for every retained origin.

Annualized revenue spans more than four orders of magnitude: the 25th and 75th percentiles are approximately \$33K and \$290K, respectively, and the observed maximum exceeds \$90M. This cross-sectional scale variation, together with short histories and heterogeneous account activity, motivates a global relative-target model rather than separate dollar-scale forecasters for each business.

A second archive contains 7,094 additional companies. For each company, one forecast origin is sampled uniformly from January--May 2025 using random seed 42. Training and evaluation companies remain disjoint, and the frozen checkpoints and scoring pipeline are applied without adaptation.

\begin{table}[t]
\centering
\caption{Company-disjoint cohorts. Origins apply the common eligibility filter; 1,060 of 1,095 test firms contribute at least one.}
\label{tab:data}
\setlength{\tabcolsep}{3.2pt}
\small
\begin{tabular}{lrr}
\toprule
Cohort & Companies & Origins \\
\midrule
Train & 5,082 & 54,836 \\
Validation & 1,086 & 11,906 \\
Test, assigned & 1,095 & -- \\
Test, evaluated & 1,060 & 11,993 \\
Later-origin & 7,094 & 7,094 \\
\bottomrule
\end{tabular}
\end{table}

\section{Accounting Graph Transformer}
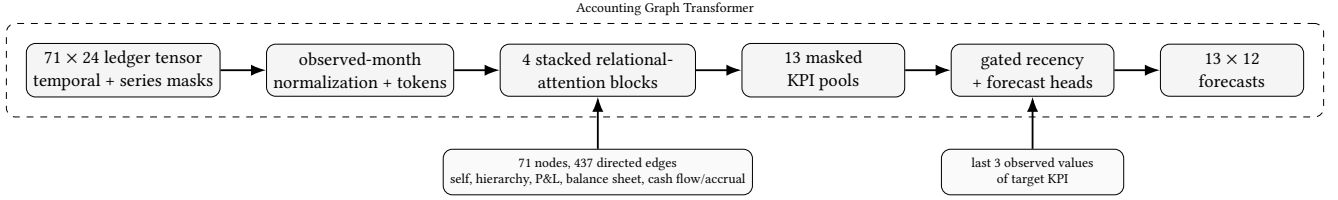
\begin{figure*}[t]
\centering
\resizebox{0.98\textwidth}{!}{%
\begin{tikzpicture}[
  node distance=7mm and 7mm,
  box/.style={draw, rounded corners, minimum height=8mm, align=center, fill=black!4, font=\small},
  smallbox/.style={draw, rounded corners, minimum height=7mm, align=center, fill=black!2, font=\scriptsize},
  arr/.style={-Latex, thick}
]
\node[box, minimum width=27mm] (input) {$71\times24$ ledger tensor\\temporal + series masks};
\node[box, minimum width=26mm, right=of input] (embed) {observed-month\\normalization + tokens};
\node[box, minimum width=30mm, right=of embed] (gat) {4 stacked relational-\\attention blocks};
\node[box, minimum width=25mm, right=of gat] (pool) {13 masked\\KPI pools};
\node[box, minimum width=25mm, right=of pool] (head) {gated recency\\+ forecast heads};
\node[box, minimum width=22mm, right=of head] (out) {$13\times12$\\forecasts};
\draw[arr] (input)--(embed); \draw[arr] (embed)--(gat); \draw[arr] (gat)--(pool); \draw[arr] (pool)--(head); \draw[arr] (head)--(out);
\node[smallbox, below=8mm of gat, minimum width=45mm] (graph) {71 nodes, 437 directed edges\\self, hierarchy, P\&L, balance sheet, cash flow/accrual};
\draw[arr] (graph)--(gat);
\node[smallbox, below=8mm of head, minimum width=28mm] (recent) {last 3 observed values\\of target KPI};
\draw[arr] (recent)--(head);
\node[draw,dashed,rounded corners,fit=(input)(embed)(gat)(pool)(head)(out),inner sep=3mm,label={[font=\scriptsize]above:Accounting Graph Transformer}] {};
\end{tikzpicture}%
}
\caption{\model{} maps a masked ledger panel to joint KPI forecasts. Sparse relation attention uses a fixed accounting graph, and each output head receives a direct three-month recency representation.}
\Description{A left-to-right pipeline maps a masked 71-by-24 ledger tensor to series tokens, applies four stacked relational-attention blocks, pools thirteen KPI-specific representations, fuses each with a three-month recency branch, and outputs thirteen twelve-month forecasts.}
\label{fig:arch}
\end{figure*}

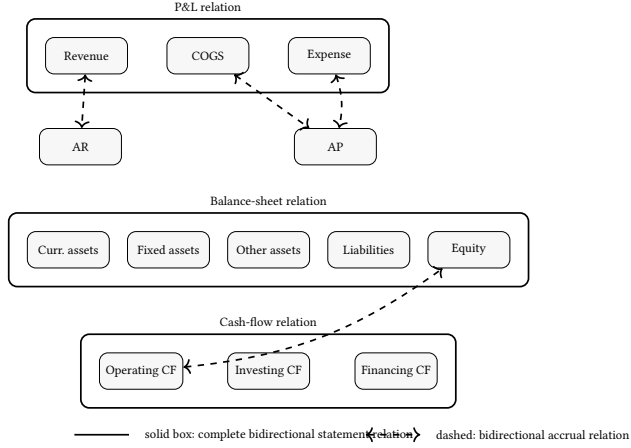
\begin{figure}[t]
\centering
\resizebox{0.98\columnwidth}{!}{%
\begin{tikzpicture}[
 n/.style={draw,rounded corners,minimum width=13.5mm,minimum height=6mm,align=center,font=\scriptsize,fill=black!3},
 group/.style={draw,rounded corners,thick,inner sep=3mm},
 accrual/.style={<->,dashed,thick},
 legend/.style={font=\scriptsize,anchor=west}
]
\node[n] (rev) at (0,2.5) {Revenue};
\node[n] (cogs) at (2.0,2.5) {COGS};
\node[n] (exp) at (4.0,2.5) {Expense};
\node[group,fit=(rev)(cogs)(exp),label={[font=\scriptsize]above:P\&L relation}] (pnl) {};

\node[n] (ar) at (-0.1,1.0) {AR};
\node[n] (ap) at (4.1,1.0) {AP};
\draw[accrual] (rev)--(ar);
\draw[accrual] (cogs)--(ap);
\draw[accrual,bend left=15] (exp) to (ap);

\node[n] (ca) at (-0.3,-0.7) {Curr. assets};
\node[n] (fa) at (1.35,-0.7) {Fixed assets};
\node[n] (oa) at (3.0,-0.7) {Other assets};
\node[n] (liab) at (4.65,-0.7) {Liabilities};
\node[n] (eq) at (6.3,-0.7) {Equity};
\node[group,fit=(ca)(fa)(oa)(liab)(eq),label={[font=\scriptsize]above:Balance-sheet relation}] (bs) {};

\node[n] (ocf) at (0.9,-2.7) {Operating CF};
\node[n] (icf) at (3.0,-2.7) {Investing CF};
\node[n] (fcf) at (5.1,-2.7) {Financing CF};
\node[group,fit=(ocf)(icf)(fcf),label={[font=\scriptsize]above:Cash-flow relation}] (cf) {};
\draw[accrual,bend right=15] (ocf) to (eq);

\draw[thick] (-0.2,-3.75)--(0.7,-3.75);
\node[legend] at (0.85,-3.75) {solid box: complete bidirectional statement relation};
\draw[accrual] (4.6,-3.75)--(5.5,-3.75);
\node[legend] at (5.65,-3.75) {dashed: bidirectional accrual relation};
\end{tikzpicture}%
}
\caption{Illustrative top-level relation types. A solid group denotes all-to-all directed attention among its members; dashed links denote operational accrual relations. Parent--child, sibling, and self-loop relations for the full 71-node graph are listed in Table~\ref{tab:edges}.}
\Description{Three solid boxes group the P and L, balance-sheet, and cash-flow variables. Dashed bidirectional links connect revenue to accounts receivable, COGS and expense to accounts payable, and operating cash flow to equity. A legend explains the two line styles.}
\label{fig:graph}
\end{figure}

\subsection{Masked series tokens}
Suppressing origin $o$, let $x_{i,\tau}$ be series $i$ at lookback position $\tau\in\{1,\ldots,24\}$, $T_{i,\tau}\in\{0,1\}$ its temporal mask, and $m_i\in\{0,1\}$ its series-availability mask. With $\Omega_i=\{\tau:T_{i,\tau}=1\}$, let $\mu_i^{\mathrm{obs}}$ and $\sigma_i^{\mathrm{obs}}$ be the mean and standard deviation over $\Omega_i$. For token width $d$,
\begin{equation}
 \hat x_{i,\tau}=T_{i,\tau}\frac{x_{i,\tau}-\mu_i^{\mathrm{obs}}}{\max(\sigma_i^{\mathrm{obs}},\epsilon)},
 \qquad h_i^{(0)}=m_i\bigl(W_p\hat x_i+e_i\bigr),
 \label{eq:embed}
\end{equation}
where $\epsilon=10^{-6}$, $W_p\in\mathbb R^{d\times24}$ is shared, and $e_i,h_i^{(0)}\in\mathbb R^d$ are the slot embedding and initial token. Masked months contribute zero to the projection, and an unavailable series cannot enter a residual path. Constant active series retain their slot embedding even when their standardized temporal component is zero.

\subsection{Accounting-relation attention}
The fixed graph contains 71 nodes and 437 directed edges in five relation types (Table~\ref{tab:edges}; Figure~\ref{fig:graph}). Hierarchy edges connect each parent to its ranked children and connect siblings within a parent. The 326 hierarchy edges comprise 116 bidirectional parent--child edges and 210 directed sibling edges: $3(6\times5)+5(4\times3)+3(4\times3)+2(4\times3)$. Statement edges connect the top-level series that participate in the P\&L, balance-sheet, and cash-flow groups. The cash-flow/accrual relation also includes bidirectional revenue--AR, COGS--AP, expense--AP, and operating-cash-flow--equity links. These latter pairs are operational priors rather than exact identities. Self-loops preserve a node's own representation in every relational block. The resulting graph uses 8.7\% of the 5,041 edges in a complete 71-node digraph before company-specific masking.

\begin{table}[t]
\centering
\caption{Fixed accounting graph. Non-self relations are directed in both directions.}
\label{tab:edges}
\setlength{\tabcolsep}{3.0pt}
\small
\begin{tabular}{lrl}
\toprule
Relation type & Edges & Construction \\
\midrule
Self & 71 & one loop per series \\
Hierarchy & 326 & parent--child and siblings \\
P\&L & 6 & revenue, COGS, expense \\
Balance sheet & 20 & assets, liabilities, equity \\
Cash flow/accrual & 14 & cash-flow components and soft links \\
\bottomrule
\end{tabular}
\end{table}

Let $\mathcal R$ be the five relation types and $\mathcal N_r(j)$ the incoming neighbors of destination node $j$ under relation $r$. For one attention head (head index omitted), with current token $h_i\in\mathbb R^d$ and head width $d_h=d/4$,
\begin{equation}
\tilde h^r_j=\sum_{i\in\mathcal{N}_r(j)}\alpha^r_{ij}W^r_Vh_i,
\quad
\alpha^r_{ij}=\operatorname{softmax}_{i\in\mathcal{N}_r(j)}
\frac{(W_Qh_j)^\top W^r_Kh_i}{\sqrt{d_h}}.
\end{equation}
Here $W_Q,W_K^r,W_V^r$ are learned query, key, and value projections, and $\alpha^r_{ij}$ is the normalized edge weight. The four head outputs are concatenated to form the relation context $\tilde h_j^r\in\mathbb R^d$. Absent sources and relation types with no valid neighbor are masked. The relation contexts are combined by
\begin{equation}
 h'_j=W_O\sum_{r\in\mathcal R}\rho_{j,r}\tilde h^r_j,
 \qquad \rho_{j,:}=\operatorname{softmax}(W_G\operatorname{LN}(h_j)),
 \label{eq:relationgate}
\end{equation}
where $W_O$ and $W_G$ are learned output and relation-gate projections, $\operatorname{LN}$ is layer normalization, and $\rho_{j,r}$ is the destination-specific weight of relation $r$. Each update is followed by a pre-normalized feed-forward network (FFN) residual block; four such blocks contextualize the tokens before pooling.

\subsection{Information flow with sparse accounts}
Account availability varies substantially across firms, so the graph is evaluated on an induced company-specific subgraph. If a child account is absent, its token is zeroed and it is removed from every attention softmax. If all incoming neighbors of relation $r$ are absent for node $j$, that relation is removed from the relation gate before renormalization. A company with no active fixed-asset children, for example, can still represent the top-level fixed-assets series and receive balance-sheet context, but inactive child slots do not contribute placeholder messages.

The hierarchy edges serve two purposes. Parent--child edges expose composition changes, while sibling edges let active subaccounts under the same parent compare their trajectories. The parent token remains directly observed and is never replaced by an aggregation of its children. This is important because the catch-all and ranked children explain composition but need not reconstruct the parent perfectly after normalization. Target-specific pooling then decides whether a forecast should rely on the parent, a child pattern, another statement item, or a combination.

This masking design also separates company age from account inactivity. The temporal mask removes unobserved leading months inside every series, whereas the series mask removes an entire channel. The distinction prevents a young company from being interpreted as a mature company with a sequence of true zeros and prevents an inactive account from contributing a learned slot embedding by itself.

\subsection{Target-specific pooling and recency}
For KPI $k$, let $q_k\in\mathbb R^d$ be a learned query. With $\log 0=-\infty$, masked pooling is
\begin{equation}
\alpha_{k,i}=\operatorname{softmax}_i(q_k^\top h_i+\log m_i),\qquad
z_k=\sum_i\alpha_{k,i}h_i,
\end{equation}
where $\alpha_{k,i}$ is the pooling weight and $z_k\in\mathbb R^d$ the KPI context. Let $\tilde x^{(3)}_k\in\mathbb R^3$ be the last three observed values of KPI $k$'s top-level series, normalized within that window. The recency path is
\begin{equation}
 s_k=W_r\tilde x^{(3)}_k,\quad a_k=\sigma(W_g[z_k;s_k]),\quad
 f_k=a_k\odot s_k+(1-a_k)\odot z_k,
 \label{eq:recency}
\end{equation}
where $W_r\in\mathbb R^{d\times3}$ and $W_g\in\mathbb R^{d\times2d}$ are learned, $\sigma$ is the sigmoid, $[\,;\,]$ denotes concatenation, and $\odot$ is elementwise multiplication. Thus $s_k$, $a_k$, and $f_k$ are the recency vector, fusion gate, and fused KPI representation. A two-layer KPI head maps $f_k$ to 12 horizons. Every company has at least 12 observed months, so this window is fully observed.

Let $\mathcal O_{\mathrm{tr}}$ be the training-origin set and $M_{o,k}\in\{0,1\}$ indicate whether KPI $k$ is evaluable at origin $o$. With Huber threshold $\delta=1$,
\begin{equation}
\mathcal{L}=\frac{1}{13}\sum_{k=1}^{13}
\frac{\sum_{o\in\mathcal O_{\mathrm{tr}}} M_{o,k}\sum_{h=1}^{12}\operatorname{Huber}_{1}(y_{o,k,h}-\hat y_{o,k,h})}
{12\sum_{o\in\mathcal O_{\mathrm{tr}}} M_{o,k}}.
\label{eq:loss}
\end{equation}
Every KPI therefore contributes equally regardless of its number of valid cells. The selected model uses width 256, four heads, four graph blocks, FFN width 512, dropout 0.1, and 5.3M parameters.

The sparse graph contains fewer than one tenth as many edges as a complete 71-node digraph. Relational attention therefore restricts both information flow and computation. The model produces all 156 KPI--horizon outputs in one forward pass and requires no company-specific fine-tuning.

\section{Experimental Design}
\subsection{Baselines and input parity}
We compare against trailing-mean and last-value forecasts; AutoETS and AutoARIMA from StatsForecast 1.6.0~\cite{nixtla2023statsforecast}; LightGBM~\cite{ke2017lightgbm}; SOFTS, TimeMixer, TiDE, iTransformer, PatchTST, NLinear, and DLinear; and Chronos-2, TimesFM 2.5, and Moirai 2.0-R. The foundation checkpoints are \path{amazon/chronos-2}, \path{google/timesfm-2.5-200m-pytorch}~\cite{google2025timesfm25}, and \path{Salesforce/moirai-2.0-R-small}. Chronos-2 and Moirai receive the 71 series jointly; TimesFM is called per series. TimesFM's regressor interface is not used because the other ledger channels are themselves unknown over the 12-month forecast horizon, rather than future-known covariates. Validation selection compared the common scaled input with model-native raw-dollar input and the predictive mean with the median; raw-dollar input and the median were selected. Chronos-2 is also fine-tuned with low-rank adaptation (LoRA), rank 8, learning rate $10^{-5}$, batch size 32, and 2,400 validation-selected steps.

All methods use the same train, validation, and test origins, observed entries, temporal and series availability, effective history length, target conversion, and scoring rules. This parity was verified at the cached-input level. No method treats an unobserved leading month as an observed zero or receives a future ledger value. Models with mask interfaces consume the masks directly; interfaces without a temporal-mask argument receive only the observed prefix. LightGBM receives the flattened histories together with availability and history-length features. Foundation models receive the equivalent raw-dollar histories required by their model-native scaling, and their outputs are converted through Eq.~\eqref{eq:target} before the common scoring step.

\begin{table}[t]
\centering
\caption{Benchmark inputs and treatment of unobserved history.}
\label{tab:baselineinfo}
\setlength{\tabcolsep}{2.3pt}
\renewcommand{\arraystretch}{1.08}
\small
\begin{tabular}{p{0.19\columnwidth}p{0.26\columnwidth}p{0.45\columnwidth}}
\toprule
Family & Forecast-time input & Availability treatment \\
\midrule
Joint neural & $71\times24$ panel & temporal and series masks \\
LightGBM & flattened 71-series panel & availability features + history length \\
Naive / stat. & target history & observed prefix only \\
Multiv. foundation & joint 71-series panel & observed prefix + supported masks \\
TimesFM & univariate calls & observed prefix; no known-future inputs \\
\bottomrule
\end{tabular}
\end{table}

Validation favored per-KPI horizon-stacked LightGBM over recursive and joint-KPI formulations. The selected model stacks the 12 horizons as rows and includes horizon as a feature. Neural baselines forecast all 13 KPIs jointly. SOFTS and TimeMixer use width 256, while the remaining neural backbones use their published default widths. We optimize \model{} with AdamW~\cite{loshchilov2019adamw} and cosine learning-rate decay. For trainable neural models, learning rate is selected from $\{10^{-3},3\times10^{-4},10^{-4}\}$, weight decay from $\{10^{-2},10^{-4}\}$, and width from $\{256,512\}$ where supported. Checkpoints are selected by validation MAE. AGT, SOFTS, TimeMixer, and LightGBM---the four strongest validation-selected trainable methods---are run with seeds 42, 123, and 7. Seed 42 was fixed in advance as the primary seed for single-checkpoint diagnostics; seeds 123 and 7 provide independent replications. The remaining benchmark rows are validation-selected point estimates. The test set is not used to select learning rate, width, stopping epoch, foundation-model point estimator, or fine-tuning length.

\begin{table}[t]
\centering
\caption{Selected AGT training configuration.}
\label{tab:training}
\setlength{\tabcolsep}{3.1pt}
\small
\begin{tabular}{lr}
\toprule
Setting & Value \\
\midrule
Token width / attention heads & 256 / 4 \\
Relational blocks / FFN width & 4 / 512 \\
Recency window / dropout & 3 months / 0.1 \\
Optimizer / learning rate & AdamW / $3\times10^{-4}$ \\
Weight decay / schedule & $10^{-4}$ / cosine \\
Batch size / maximum epochs & 256 / 80 \\
Early-stopping patience & 15 epochs \\
Huber threshold & 1.0 \\
Parameters & 5.3M \\
\bottomrule
\end{tabular}
\end{table}

\subsection{Inference and ablations}
Across-seed mean and standard deviation characterize optimization variability for AGT, LightGBM, TimeMixer, and SOFTS. Paired company-clustered intervals use their pre-specified seed-42 checkpoints, so every baseline comparison shares the same fitted AGT reference. Each bootstrap draw resamples companies with replacement and recomputes the sample-weighted KPI-macro MAE, preserving covariance among KPIs and overlapping origins.

The final-architecture ablations use seed 42, identical optimization settings, and the same validation-selection rule. They remove graph attention, remove the recency path, or replace the accounting graph with a degree-matched random topology. For the random control, non-self-loop destinations are permuted while sources remain fixed, preserving source out-degree and destination in-degree while destroying the accounting pairings.

\section{Results}
\subsection{Main benchmark}
\begin{table}[t]
\centering
\caption{Sample-weighted KPI-macro test MAE. AGT, LightGBM, TimeMixer, and SOFTS show mean $\pm$ standard deviation over three independent seeds; other entries are validation-selected point estimates. ZS denotes zero-shot and FT fine-tuned. Best in each category is bold; lower is better.}
\label{tab:main}
\setlength{\tabcolsep}{4.0pt}
\renewcommand{\arraystretch}{1.06}
\small
\begin{tabular}{llr}
\toprule
Category & Model & MAE \\
\midrule
Naive & \textbf{Trailing mean} & \textbf{0.9020} \\
      & Last value & 1.0880 \\
Statistical & \textbf{AutoETS} & \textbf{1.0730} \\
            & AutoARIMA & 1.1200 \\
Tree & \textbf{LightGBM} & $\mathbf{0.7378\pm0.0014}$ \\
Neural & SOFTS & $0.7560\pm0.0041$ \\
       & \textbf{TimeMixer} & $\mathbf{0.7523\pm0.0021}$ \\
       & TiDE & 0.8220 \\
       & iTransformer & 0.8470 \\
       & PatchTST & 0.8970 \\
       & NLinear & 1.1780 \\
       & DLinear & 1.2110 \\
Foundation & Chronos-2 ZS & 0.8320 \\
           & \textbf{Chronos-2 FT} & \textbf{0.8010} \\
           & TimesFM 2.5 & 0.8640 \\
           & Moirai 2.0-R & 0.9110 \\
\midrule
Ours & \textbf{AGT} & $\mathbf{0.6990\pm0.0013}$ \\
\bottomrule
\end{tabular}
\end{table}

\begin{table}[t]
\centering
\caption{Independent-seed test MAE for the four strongest trainable methods. LGBM denotes LightGBM and TMixer TimeMixer; SD is standard deviation.}
\label{tab:seeds}
\setlength{\tabcolsep}{1.6pt}
\footnotesize
\begin{tabular}{rrrrr}
\toprule
Seed & AGT & LGBM & TMixer & SOFTS \\
\midrule
42 & 0.6979 & 0.7374 & 0.7521 & 0.7566 \\
123 & 0.6987 & 0.7393 & 0.7503 & 0.7516 \\
7 & 0.7004 & 0.7366 & 0.7544 & 0.7598 \\
\midrule
Mean $\pm$ SD & $0.6990\pm0.0013$ & $0.7378\pm0.0014$ & $0.7523\pm0.0021$ & $0.7560\pm0.0041$ \\
\bottomrule
\end{tabular}
\end{table}

Across three seeds, AGT is the lowest-MAE method at every seed and improves on the mean LightGBM score by 5.3\%, TimeMixer by 7.1\%, and SOFTS by 7.5\%, using each baseline as denominator. At the pre-specified seed-42 checkpoint, the paired company-clustered differences are 0.0395 against LightGBM (95\% CI $[0.0350,0.0439]$), 0.0542 against TimeMixer ($[0.0500,0.0583]$), and 0.0587 against SOFTS ($[0.0543,0.0634]$). All three intervals exclude zero, and AGT remains the lowest-MAE method at seeds 123 and 7.

\begin{table}[t]
\centering
\caption{Paired company-clustered seed-42 differences against the three strongest baselines. Positive values favor AGT.}
\label{tab:paired}
\setlength{\tabcolsep}{3.0pt}
\small
\begin{tabular}{lrr}
\toprule
Comparison & Difference & 95\% CI \\
\midrule
LightGBM$-$AGT & 0.0395 & [0.0350, 0.0439] \\
TimeMixer$-$AGT & 0.0542 & [0.0500, 0.0583] \\
SOFTS$-$AGT & 0.0587 & [0.0543, 0.0634] \\
\bottomrule
\end{tabular}
\end{table}

The relative result is stable under company balancing. Table~\ref{tab:estimand} shows that the SOFTS-minus-AGT gap is 0.0587 with sample weighting and 0.0583 when each company receives equal weight. The corresponding AGT advantages over LightGBM and TimeMixer are also stable.

\begin{table}[t]
\centering
\caption{Matched seed-42 point estimates under sample and company-balanced weighting.}
\label{tab:estimand}
\setlength{\tabcolsep}{3.2pt}
\small
\begin{tabular}{lrr}
\toprule
Model & Sample-weighted & Company-balanced \\
\midrule
AGT & \textbf{0.6979} & \textbf{0.7268} \\
LightGBM & 0.7374 & 0.7668 \\
TimeMixer & 0.7521 & 0.7834 \\
SOFTS & 0.7566 & 0.7851 \\
\bottomrule
\end{tabular}
\end{table}

\subsection{Comparison by model class}
The benchmark separates three sources of forecasting capability. LightGBM receives the complete flattened ledger panel, explicit masks, history length, and horizon features; its three-seed mean of 0.7378 shows that the task contains substantial predictive signal even without a sequence encoder. At seed 42, AGT's paired advantage of 0.0395 shows that the structured joint model improves further on this strong feature-based reference.

SOFTS and TimeMixer are the closest generic neural competitors. Both process the joint multivariate history and are tuned on the same validation split, but neither receives a fixed accounting neighborhood. Their three-seed means are 0.7560 and 0.7523, respectively, showing that the gap is not specific to one baseline implementation. AGT's advantage over both exceeds 0.05 MAE and remains visible under company-balanced weighting.

The foundation-model results show a different pattern. Chronos-2 fine-tuning improves over its zero-shot checkpoint (0.801 versus 0.832), while TimesFM and Moirai remain above the trained task-specific models. General pretraining therefore transfers useful temporal knowledge, but task-specific learning remains important in a sparse ledger panel. Chronos-2 and Moirai also receive all 71 series; AGT additionally conditions cross-series exchange on the fixed financial graph while optimizing the 13 targets jointly.

\subsection{KPI-level performance}
Table~\ref{tab:perkpi} reports the seed-42 checkpoints for AGT and the three strongest baselines. AGT is best on all 13 targets. The improvement spans the income statement, all five balance-sheet outputs, all three cash-flow components, and AR/AP, showing that the aggregate result is not driven by a single volatile KPI. Averaged within statement families, AGT obtains 0.544 on income-statement KPIs, 0.409 on balance-sheet KPIs, 1.396 on cash-flow KPIs, and 0.604 on working-capital KPIs. The corresponding SOFTS values are 0.572, 0.465, 1.490, and 0.661.

Operating cash flow is the hardest target for every method. AGT improves operating-cash-flow MAE by 0.079 versus SOFTS, while its largest absolute gain is on other assets (0.140). Revenue is the closest comparison, where AGT still leads all three baselines. The gains therefore extend from the most difficult output to the closest baseline comparison.

\begin{table*}[t]
\centering
\caption{Per-KPI sample-weighted MAE for the matched seed-42 checkpoints. Bold is best in each row.}
\label{tab:perkpi}
\setlength{\tabcolsep}{5.5pt}
\small
\begin{tabular}{lrrrr}
\toprule
KPI & AGT & SOFTS & LightGBM & TimeMixer \\
\midrule
Revenue & \textbf{0.5256} & 0.5295 & 0.5300 & 0.5384 \\
Expense & \textbf{0.4591} & 0.4828 & 0.4773 & 0.4841 \\
COGS & \textbf{0.6482} & 0.7041 & 0.6689 & 0.6807 \\
Current assets & \textbf{0.4190} & 0.4442 & 0.4222 & 0.4391 \\
Fixed assets & \textbf{0.3392} & 0.4013 & 0.3477 & 0.3844 \\
Other assets & \textbf{0.2680} & 0.4078 & 0.2891 & 0.3263 \\
Liabilities & \textbf{0.4393} & 0.4693 & 0.4487 & 0.4655 \\
Equity & \textbf{0.5776} & 0.6036 & 0.5906 & 0.6050 \\
Operating cash flow & \textbf{2.1119} & 2.1905 & 2.2466 & 2.2287 \\
Investing cash flow & \textbf{0.9813} & 1.1060 & 1.1439 & 1.1377 \\
Financing cash flow & \textbf{1.0947} & 1.1737 & 1.1818 & 1.2081 \\
AR total & \textbf{0.6014} & 0.6585 & 0.6180 & 0.6341 \\
AP total & \textbf{0.6073} & 0.6640 & 0.6212 & 0.6450 \\
\midrule
Overall & \textbf{0.6979} & 0.7566 & 0.7374 & 0.7521 \\
\bottomrule
\end{tabular}
\end{table*}

\begin{table}[t]
\centering
\caption{Seed-42 MAE averaged within financial-statement families.}
\label{tab:families}
\setlength{\tabcolsep}{2.7pt}
\renewcommand{\arraystretch}{1.06}
\small
\begin{tabular}{lrrrr}
\toprule
Family & AGT & SOFTS & LightGBM & TimeMixer \\
\midrule
Income statement & \textbf{0.5443} & 0.5721 & 0.5587 & 0.5677 \\
Balance sheet & \textbf{0.4086} & 0.4652 & 0.4197 & 0.4441 \\
Cash flow & \textbf{1.3960} & 1.4901 & 1.5241 & 1.5248 \\
Working capital & \textbf{0.6043} & 0.6613 & 0.6196 & 0.6396 \\
\bottomrule
\end{tabular}
\end{table}

Across financial-statement families, AGT's largest proportional gain over SOFTS is on the balance-sheet KPIs (12.2\%). Its largest family-level gain over LightGBM is on cash flow (8.4\%). The family view complements the 13-row table by showing that the improvement spans statement stocks, statement flows, and working capital.

\subsection{Final-architecture ablations}
Table~\ref{tab:ablation} evaluates the components of the submitted architecture. Removing graph attention causes the largest degradation, with a company-clustered paired increase of 0.0141 and 95\% CI $[0.0112,0.0171]$. Replacing the accounting topology with a degree-matched random graph gives a paired increase of 0.0063, and removing the recency path gives 0.0053. Each ablation is also worse than AGT on validation.

\begin{table}[t]
\centering
\caption{Seed-42 ablations of final AGT. Validation (Val) and test are sample-weighted MAE; $\Delta=\mathrm{MAE}_{\mathrm{ablation}}-\mathrm{MAE}_{\mathrm{AGT}}$, with a company-clustered 95\% CI.}
\label{tab:ablation}
\setlength{\tabcolsep}{2.6pt}
\small
\begin{tabular}{lrrr}
\toprule
Configuration & Val & Test & $\Delta$ [95\% CI] \\
\midrule
AGT & \textbf{0.6885} & \textbf{0.6979} & -- \\
No graph attention & 0.7047 & 0.7119 & 0.0141 [0.0112, 0.0171] \\
Random graph & 0.6936 & 0.7039 & 0.0063 [0.0039, 0.0088] \\
No recency path & 0.6939 & 0.7030 & 0.0053 [0.0033, 0.0072] \\
\bottomrule
\end{tabular}
\end{table}

All ablations share the same token encoder, masks, KPI pooling, output heads, training budget, and validation rule. The random control preserves graph capacity while permuting the accounting pairings; the no-graph variant removes the relational stage. AGT outperforms both, separating the contribution of accounting topology from the broader contribution of relational processing.

At the sample-weighted point estimate, random connectivity recovers part of the no-graph gap, and the accounting topology improves further. The validation values follow the same direction: AGT is 0.6885, compared with 0.6936 for the random graph, 0.6939 without recency, and 0.7047 without graph attention.

\subsection{Later-origin transfer}
Table~\ref{tab:later} evaluates one uniformly sampled January--May 2025 origin for each of 7,094 additional companies (random seed 42). AGT remains the strongest model, with 0.7548 MAE compared with 0.7694 for SOFTS, 0.787 for TimeMixer, and 0.809 for LightGBM, an absolute improvement of 0.0146 over the closest baseline. The ordering extends to a distinct company population and origin range. Because each company contributes exactly one origin, sample-weighted and company-balanced MAE coincide on this cohort.

AGT requires no adaptation or company-specific fine-tuning between the primary and later-origin evaluations.

\begin{table}[t]
\centering
\caption{Later-origin transfer for the seed-42 checkpoints: 7,094 unseen companies, one uniformly sampled January--May 2025 origin per company.}
\label{tab:later}
\setlength{\tabcolsep}{4.2pt}
\small
\begin{tabular}{lr}
\toprule
Model & MAE \\
\midrule
AGT & \textbf{0.7548} \\
SOFTS & 0.7694 \\
TimeMixer & 0.7870 \\
LightGBM & 0.8090 \\

\bottomrule
\end{tabular}
\end{table}

\section{Discussion and Conclusion}
The component results align with AGT's design. Relational blocks restrict repeated cross-series updates to financially plausible neighborhoods, while the recency path supplies a direct local signal from the target KPI. Removing relational attention causes the largest degradation in the final architecture, and replacing the accounting graph with a degree-matched random topology also reduces accuracy. The graph and recency path therefore provide complementary cross-series and local information.

The improvement is broad across the financial system. AGT leads the three strongest baselines on every individual KPI, spanning the income statement, all five balance-sheet outputs, the three cash-flow components, and AR/AP. The per-KPI results support output-specific review thresholds rather than one global tolerance for all 13 forecasts.

The main comparison is stable across optimization seeds and company weighting. AGT's three test MAEs lie between 0.6979 and 0.7004, while every matched SOFTS run is above 0.751. Giving each company equal influence changes the absolute scores but leaves the SOFTS-minus-AGT gap nearly unchanged (0.0583 versus 0.0587 with sample weighting). On the later-origin cohort, the same checkpoint is applied without adaptation to 7,094 additional companies and retains the best MAE, supporting one panel model rather than separate firm-level fits.

The benchmark spans distinct alternatives rather than variants of one backbone. LightGBM receives the complete ledger panel and explicit availability features; TimeMixer and SOFTS learn generic joint representations; Chronos-2 and Moirai receive all 71 series; and TimesFM uses its native per-series interface. AGT's lead across these classes is therefore not tied to one baseline family or one treatment of masked history.

Ranked child slots make the input representation usable across heterogeneous charts of accounts. Each slot retains stable parent and activity-rank semantics, while catch-all channels preserve the long tail of company-specific accounts. The temporal and series masks distinguish a young business from an inactive account, and the fixed graph supplies a common interaction pattern across firms despite differences in account activity.

A single 5.3M-parameter checkpoint produces all 156 KPI--horizon forecasts in one pass. The mean-relative outputs can be inverted exactly to dollar scenarios through Eq.~\eqref{eq:invert}, allowing planning, liquidity, working-capital, and balance-sheet views to share one forecast origin and encoded ledger state. This joint output is operationally different from maintaining separate firm--KPI models: it produces internally aligned scenarios under one data cutoff and can be refreshed for a new company without refitting. In consequential workflows, the trajectories remain decision-support inputs to current business context, uncertainty assessment, analyst review, and policy controls.

In summary, AGT combines a fixed accounting-relation graph with target-specific recency conditioning for joint forecasting from sparse small-business ledgers. It achieves $0.6990\pm0.0013$ MAE over three independent seeds, wins on all 13 KPIs in the matched comparison, and remains first on a later-origin cohort of 7,094 unseen companies. The final-architecture ablations show that accounting-structured attention and recent target history each contribute to predictive accuracy, establishing fixed financial structure as a practical inductive bias for multi-KPI forecasting from short monthly histories.

\end{document}